\documentclass[a4paper,twoside]{article}

\usepackage{epsfig}
\usepackage{subcaption}
\usepackage{calc}
\usepackage{amssymb}
\usepackage{amstext}
\usepackage{amsmath}
\usepackage{amsthm}
\usepackage{multicol}
\usepackage{pslatex}
\usepackage{apalike}
\usepackage[bottom]{footmisc}
\usepackage{graphicx}
\usepackage{import}
\usepackage{hyperref}
\usepackage{algorithm}
\usepackage{algpseudocode}
\usepackage{accents}
\usepackage{xcolor}

\usepackage{SCITEPRESS}     

\algnewcommand\algorithmicinput{\textbf{Input:}}
\algnewcommand\Input{\item[\algorithmicinput]}

\algnewcommand\algorithmicresult{\textbf{Result:}}
\algnewcommand\Result{\item[\algorithmicresult]}

\begin{document}

\title{MA-VAE: Multi-head Attention-based Variational Autoencoder Approach for Anomaly Detection in Multivariate Time-series Applied to Automotive Endurance Powertrain Testing}

\author{\authorname{
Lucas Correia\sup{1}\sup{2},  
Jan-Christoph Goos\sup{1}, 
Philipp Klein\sup{1}, 
Thomas Bäck\sup{2} and 
Anna V. Kononova\sup{2}}
\affiliation{\sup{1}Mercedes-Benz AG, Stuttgart, Germany}
\affiliation{\sup{2}Leiden University, Leiden, The Netherlands}
\email{lucas.correia@mercedes-benz.com}
}


\abstract{A clear need for automatic anomaly detection applied to automotive testing has emerged as more and more attention is paid to the data recorded and manual evaluation by humans reaches its capacity. Such real-world data is massive, diverse, multivariate and temporal in nature, therefore requiring modelling of the testee behaviour. We propose a variational autoencoder with multi-head attention (MA-VAE), which, when trained on unlabelled data, not only provides very few false positives but also manages to detect the majority of the anomalies presented. In addition to that, the approach offers a novel way to avoid the bypass phenomenon, an undesirable behaviour investigated in literature. Lastly, the approach also introduces a new method to remap individual windows to a continuous time series. The results are presented in the context of a real-world industrial data set and several experiments are undertaken to further investigate certain aspects of the proposed model. When configured properly, it is 9\% of the time wrong when an anomaly is flagged and discovers 67\% of the anomalies present. Also, MA-VAE has the potential to perform well with only a fraction of the training and validation subset, however, to extract it, a more sophisticated threshold estimation method is required.}

\onecolumn \maketitle \normalsize \setcounter{footnote}{0} \vfill

\section{Introduction}\label{sec:introduction}
Powertrain testing is an integral part of the wider automotive powertrain development and is undertaken at different stages of development. Each of these stages is composed of many integration levels. These integration levels range from powertrain sub-component testing, such as the electric drive unit (EDU) controller or high-voltage battery (HVB) management system, to whole vehicle powertrain testing. Each of these has its special type of controlled environment, called a test bench. The use-case in this paper is on an endurance powertrain test bench, where the EDU and HVB on their own are tested under different conditions and loads for longer periods to simulate wear over time. Given the costly maintenance and upkeep costs of such test benches, it is desirable to keep downtime at a minimum and to avoid faulty measurements. Also, it is desirable to detect problems early to prevent damage to the testee. Given that evaluation is done manually by inspection, it is not feasible to analyse every single measurement, also evaluation tends to be delayed, only being undertaken days after the measurement is recorded, hence there is a clear need for automatic, fast and unsupervised evaluation methodology which can flag anomalous measurements before the next measurement is started.

To achieve this, we propose a multi-head attention variational autoencoder (MA-VAE). MA-VAE consists of a bidirectional long short-term memory (BiLSTM) variational autoencoder architecture that maps a time-series window into a temporal latent distribution \cite{park_multimodal_2018} \cite{su_robust_2019}. Also, a multi-head attention (MA) mechanism is added to further enhance the sampled latent matrix before it is passed on to the decoder. As shown in the ablation study, this approach avoids the so-called bypassed phenomenon \cite{bahuleyan_variational_2018}, which is the first contribution. Furthermore, this paper offers a unique methodology for the reverse-window process. It is used for remapping the fixed-length windows the model is trained on to continuous variable-length sequences.

This paper is structured as follows: First, a short background is provided in Section \ref{sec:background} on the powertrain testing methodology specific to this use case, as well as the theory behind VAE and MA mechanisms. Then, related work in variational autoencoder-based time-series anomaly detection is presented in Section \ref{sec:related_work}, followed by an in-depth introduction of the real-world data set and the approach we propose in Section \ref{sec:proposed_approach}. Then, several experiments testing different aspects of the proposed method are conducted and discussed in Section \ref{sec:results}, along with the final results. Finally, conclusions from this work are drawn and an outlook into future work is provided in Section \ref{sec:conclusion}. The source code for the data pre-processing, model training as well as evaluation can be found under \url{https://github.com/lcs-crr/MA-VAE}.

\section{Background}\label{sec:background}
\subsection{Real-world Application}
During endurance testing a portfolio of different driving cycles is run, where a cycle is a standardised driving pattern, which enables repeatability of measurements. For this type of testing the portfolio consists exclusively of proprietary cycles, which differ from the public cycles used, for example, for vehicle fuel/energy consumption certification like the New European Driving Cycle (NEDC) or the Worldwide Harmonised Light Vehicles Test Cycle (WLTC). The reason why proprietary cycles are used for endurance runs is that they allow for more extensive loading of the powertrain. 

Given the presence of a battery in the testee, some time has to be dedicated to battery soaking (sitting idle) and charging. These procedures are also standardised using cycles, although, for the intents and purposes of this paper, they are omitted. What is left are the eight dynamic driving cycles representing short, long, fast, slow and dynamic trips ranging from $5$ to $30$ minutes. There are multiple versions of the same cycle, which mostly differ in starting conditions such as state-of-charge (SoC) and temperature of the battery.

On powertrain test benches, there are several control methods to ensure the testee maintains the given driving cycle. In this particular test bench, the regulation is done by the acceleration pedal and the EDU revolutions-per-minute (rpm), which is nothing more than a non-SI version of the angular velocity.

\subsection{Data Set}
This real-world data set consists of 3385 normal measurement files, each of which contains hundreds of (mostly redundant or empty) channels. A measurement is considered normal when the testee behaviour conforms to the norm. For this work, a list of $d_{\textbf{X}}=13$ channels was hand-picked in consultation with the test bench engineers to choose a reasonable and representative number of channels. This list includes the vehicle speed, EDU torque, current, voltage, rotor temperature and stator temperature, left and right wheel shaft torque, HVB current, voltage, temperature and SoC and inverter temperature. Given that some channels (such as torque) are sampled much faster than others (like temperature and SoC), a common sampling rate of $2$Hz is chosen. Channels sampled slower than $2$Hz are linearly interpolated, which is seen as permissible due to the lower amplitude resolution of those channels. Channels sampled faster than $2$Hz are passed through a low-pass filter with a cut-off frequency of $1$Hz and then resampled to $2$Hz, as is consistent with the Whittaker–Nyquist–Shannon theorem \cite{shannon_communication_1949}. Then the driving cycles are z-score normalised, i.e. transformed such that the mean for each channel lies at $0$ and the standard deviation at $1$. Lastly, the driving cycles (generally referred to as \textit{sequences} in this paper) are windowed to create a set of fixed-length sub-sequences, or \textit{windows}. First, each channel is auto-correlated to obtain the number of lags of the slowest dynamic process present in the signal. Then, the window size $W$ is set as the smallest power of two larger than the longest lag, in this case, $W=256$ time steps or $128$ seconds. Each window overlaps their preceding and succeeding windows by half a window, i.e. the shift between windows is $W/2=128$ time steps, in order to reduce computational load compared to a shift of one time step.

Due to the absence of labelled anomalies in the dataset, realistic anomalous events are intentionally simulated and recorded following the advice of test bench engineers. To this end, five anomaly types were recorded. In the first type, the virtual wheel diameter is changed, such that the resulting vehicle speed deviates from the norm. The wheel diameter is a parameter as resistances are connected to the shafts rather than actual wheels. The second type of anomaly involves changing the driving mode from comfort to sport, which leads to a higher HVB SoC drop over the cycle and a different torque response. In the third anomaly, the recuperation level is turned from maximum to zero, hence the minimum EDU torque is always non-negative and the HVB SoC experiences a higher drop in SoC. In the case of the fourth anomaly, the HVB is swapped for a battery simulator, where the HVB voltage behaviour deviates from a real battery. The inverter and EDU share a cooling loop, whose cooling capacity is reduced at the beginning or middle of the cycle, leading to higher EDU rotor, EDU stator and inverter temperatures than normal. Every anomaly type is recorded during every cycle at least once, leading to $60$ anomalous driving cycles that are all used as the anomalous subset of the test set.

\begin{figure}[!t]
    \centering
    \includegraphics[width=0.4\textwidth]{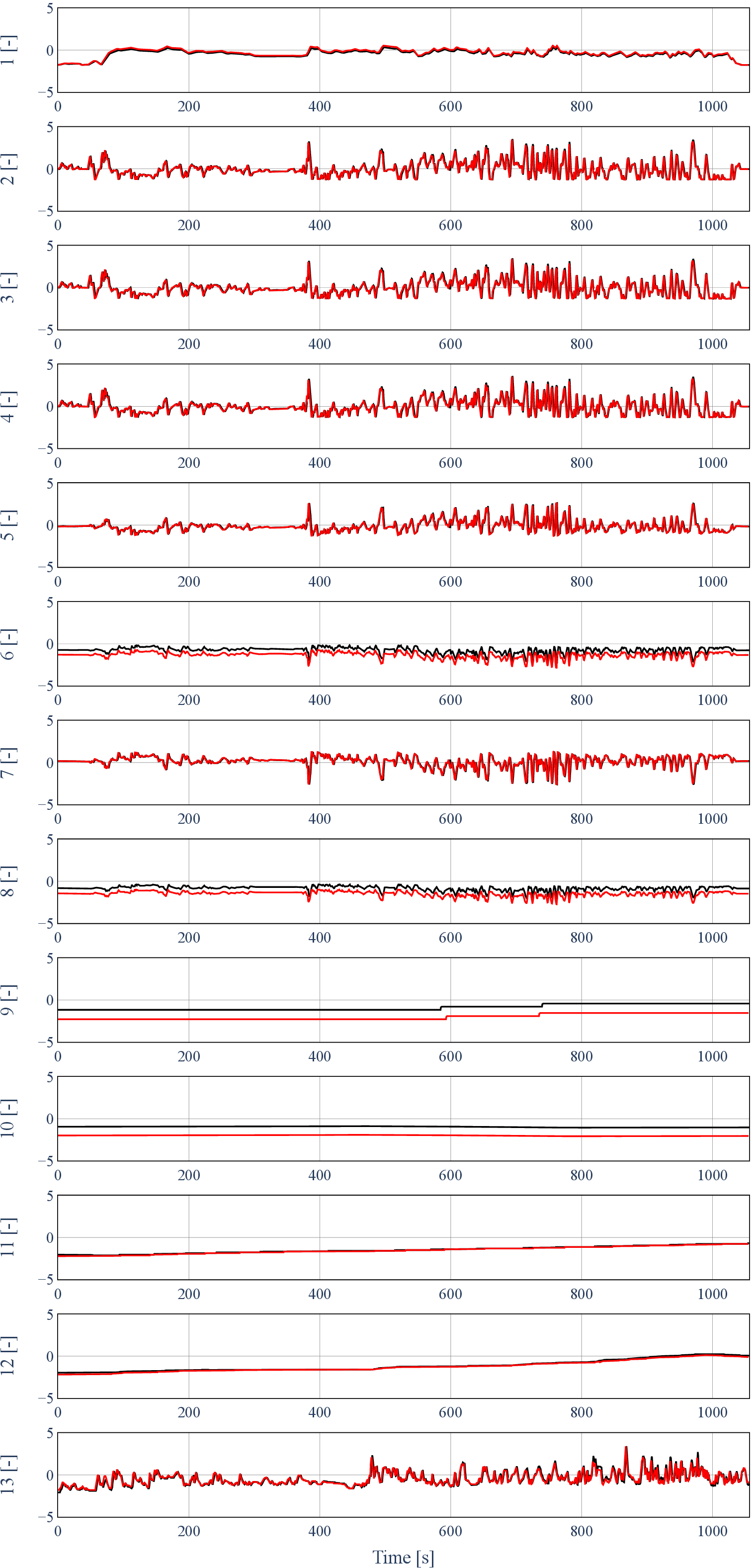}
    \caption{Features of a normal (black) and an anomalous (red) cycle plotted with respect to time. The anomalous cycle plotted represents a scenario where the wheel diameter has not been set correctly. The amplitude axis is z-score normalised to comply with confidentiality guidelines.}
    \label{fig:data_plot}
\end{figure}

A plot of one normal and one wheel-diameter anomalous cycle is shown in Figure \ref{fig:data_plot}. Due to the long channel names, the plot only shows the channel indices, a table containing the legend is shown in Table \ref{tab:legend} for context. Visual inspection may suggest that the red plot is anomalous, since the EDU and HVB voltage, temperature and state of charge deviate from the black plot. This deviation is to be expected because they depend on how charged the battery is and on how much the battery is used previous to the current cycle. In the case of this anomaly, the only channel that demonstrates anomalous behaviour is the vehicle speed, since:
\begin{equation}
    v_{\text{vehicle}}=r\times\omega
\end{equation}
where $r$ is the wheel radius and $\omega$ the angular velocity. Evidently, the anomalous behaviour is most visible at higher speeds.

\begin{table}[!b]
    \caption{Legend for the channel names in Figure \ref{fig:data_plot}.}
    \label{tab:legend}
    \centering
    \begin{tabular}{cl}
    \hline
    No. & Name                      \\ \hline \hline
    1   & Vehicle Speed             \\
    2   & EDU Torque                \\
    3   & Left Axle Torque          \\
    4   & Right Axle Torque         \\
    5   & EDU Current               \\
    6   & EDU Voltage               \\
    7   & HVB Current               \\
    8   & HVB Voltage               \\
    9   & HVB Temperature           \\
    10  & HVB State of Charge       \\
    11  & EDU Rotor Temperature     \\
    12  & EDU Stator Temperature    \\
    13  & Inverter Temperature      \\ \hline
    \end{tabular}
\end{table}
In an operative environment, it is desirable to find out whether the previously recorded sequence had any problems to analyse before the next measurement is recorded. Also, a model that performs as well as possible with as little data as possible translates to faster deployment. \textit{Good} performance is indicated by a model that can detect as many anomalies as possible and rarely labels normal measurements wrongly. To investigate the required training subset size of the model, it is trained with $1$h, $8$h, $64$h, and $512$h worth of dynamic testing data, which corresponds to the first $6$, $44$, $348$, and $2785$ driving cycles, respectively. The results are also presented in Section \ref{sec:results}. In each of the above-mentioned cases, the training subset is further split into a training ($80\%$) and a validation ($20\%$) subsets. Both the training and validation subsets are batched to sets of 512 windows.
Given the anomalous subset size of $60$ driving cycles, $600$ normal driving cycles recorded after the ones in the training subset are chosen to make up the normal test subset. This would imply that 9\% of measurements at the test bench are anomalous, however in reality this value is estimated to be much lower. This amount of anomalous data in relation to normal data is used as it approximately matches the anomaly ratio in public data sets and because the data set is not large enough to create a larger normal test subset. 

\subsection{Variational Autoencoders}
The variational autoencoder \cite{kingma_auto-encoding_2014}\cite{rezende_stochastic_2014} is a generative model that structurally resembles an autoencoder, but is theoretically derived from variational Bayesian statistics. As opposed to the regular deterministic autoencoder, the VAE uses the evidence lower bound (ELBO), which is a lower bound approximation of the so-called log evidence $\log p_\theta(\textbf{X})$, as its objective function. The ELBO, Equation \ref{eq:ELBO}, can be expressed as the reconstruction log-likelihood and the negative Kullback-Leibler Divergence ($D_{\text{KL}}$) between the approximate posterior $q_{\phi}(\textbf{Z}\vert \textbf{X})$ and the prior $p_{\theta}(\textbf{Z})$, which is typically assumed to be a Gaussian distribution \cite{goodfellow_deep_2016}.
\begin{equation}
    \begin{split}
        \mathcal{L}_{\theta, \phi}(\textbf{X}) &= \mathbb{E}_{\textbf{Z} \sim q_\phi(\textbf{Z}|\textbf{X})}  \left[ \log p_\theta(\textbf{X} | \textbf{Z}) \right] \\
        & - D_{\text{KL}}(q_\phi(\textbf{Z} | \textbf{X}) || p_\theta(\textbf{Z}))  \\
    \end{split}
    \label{eq:ELBO}
\end{equation}
where $\textbf{Z} \in \mathbb{R}^{W \times d_{\textbf{Z}}}$ is the sampled latent matrix and $\textbf{X} \in \mathbb{R}^{W \times d_{\textbf{X}}}$ is the input window. $W$ refers to the window length, whereas $d_{\textbf{X}}$ and $d_{\textbf{Z}}$ refer to the input window and latent matrix dimensionality, respectively. Gradient-based optimisation minimises an objective function and the goal is the maximisation of the ELBO, hence the final loss function is defined as the negative of Equation \ref{eq:ELBO}, shown in Equation \ref{eq:MA-VAE_loss}.
\begin{equation}
    \mathcal{L}_{\text{VAE}} = -\mathcal{L}_{\theta, \phi}(\textbf{X})
    \label{eq:MA-VAE_loss}
\end{equation}

Finally, to enable the backpropagation through the otherwise intractable gradient of the ELBO, the \textit{reparametrisation trick} \cite{kingma_auto-encoding_2014} is applied, shown in Equation \ref{eq:reparametrisation}.
\begin{equation}
    \textbf{Z} = \boldsymbol{\mu}_\textbf{Z} + \epsilon \cdot \boldsymbol{\sigma}_\textbf{Z}
    \label{eq:reparametrisation}
    \end{equation} 
where $\epsilon \sim \mathcal{N}(0, 1)$ and $(\boldsymbol{\mu}_\textbf{Z}, \log\boldsymbol{\sigma}_\textbf{Z}^2) = q_{\phi}(\textbf{X})$.

\subsection{Multi-head Attention Mechanism}
To simplify the explanation of MA as employed in this work, multi-head self-attention (MS) will be explained instead with the small difference between MA and MS being pointed out at the end.  

MS consists of two different concepts: self-attention and its multi-head extension. Self-attention is nothing more than scaled dot-product attention \cite{vaswani_attention_2017} where the key, query and value are the same. The scaled dot-product attention score is the softmax \cite{soulie_probabilistic_1990} of the product between query matrix $\textbf{Q}$ and key matrix $\textbf{K}$ which is scaled by $\sqrt{d_\textbf{K}}$. The product between the attention score and the value matrix $\textbf{V}$ yields the context matrix $\textbf{C}$, as shown in Equation \ref{eq:dot_product_attention}.
\begin{equation}
    \textbf{C} = \mathrm{Softmax}\left(\frac{\textbf{Q}\textbf{K}^T}{\sqrt{d_\textbf{K}}}\right)\textbf{V}
    \label{eq:dot_product_attention}
\end{equation}
Compared to recurrent or convolutional layers, self-attention offers a variety of benefits, such as the reduction of computational complexity, as well as an increased amount of operations that can be parallelised. \cite{vaswani_attention_2017}. Also, self-attention inherits an advantage over Bahdanau-style attention \cite{bahdanau_neural_2015} from the underlying scaled dot-product attention mechanism: it can run efficiently in matrix multiplication manner \cite{vaswani_attention_2017}.  

Multi-head self-attention then allows the attention model to attend to different representation subspaces \cite{vaswani_attention_2017}, in addition to learning useful projections rather than it being a stateless transformation \cite{chollet_deep_2021}. This is achieved using weight matrices $\textbf{W}^Q_i$, $\textbf{W}^K_i$, $\textbf{W}^V_i$, which contain trainable parameters and are unique for each head $i$, as shown in Equation\ref{eq:multi_self_attention}.
\begin{equation}
    \textbf{Q}_i = \textbf{Q}\textbf{W}^Q_i \quad \textbf{K}_i = \textbf{K}\textbf{W}^K_i \quad  \textbf{V}_i = \textbf{V}\textbf{W}^V_i
    \label{eq:multi_self_attention}
\end{equation}
Once the query, key and value matrices are linearly transformed via the weight matrices, the context matrix $\textbf{C}_i$ for each head $i$ is computed using Equation \ref{eq:multi_dot_product_attention}.
\begin{equation}
    \textbf{C}_i = \mathrm{Softmax}\left(\frac{\textbf{Q}_i\textbf{K}_i^T}{\sqrt{d_\textbf{K}}}\right)\textbf{V}_i
    \label{eq:multi_dot_product_attention}
\end{equation}
Then, for $h$ heads, the different context matrices are concatenated and linearly transformed again via the weight matrix $\textbf{W}^O$, resulting in the multi-head context matrix $\textbf{C} \in \mathbb{R}^{W \times d_{\textbf{Z}}}$, Equation \ref{eq:concat_multi_dot_product_attention}.
\begin{equation}
    \textbf{C} = [\textbf{C}_1, ..., \textbf{C}_h]\textbf{W}^{O}
    \label{eq:concat_multi_dot_product_attention}
\end{equation}
The underlying mechanism of MA is identical to MS, with the only difference being that \textbf{K} $=$ \textbf{Q} $\not =$ \textbf{V}. The benefit of this alteration is discussed in Section \ref{sec:proposed_approach}.

\section{Related Work}\label{sec:related_work}
MA-VAE belongs to the so-called generative model class, which encompasses both variational autoencoders, as well as generative adversarial networks. This section focuses solely on the work on VAE proposed in the context of time-series anomaly detection.

In time-series anomaly detection literature, the only other model that uses the combination of a VAE and an attention mechanism is by \cite{pereira_unsupervised_2018}.  For the purpose of our paper, it is named VS-VAE. Their approach consists of a BiLSTM encoder and decoder, where, for an input window of length W, the $t=W$ encoder hidden states of each direction are passed on to the variational self-attention (VS) mechanism \cite{bahuleyan_variational_2018}. The resulting context vector is then concatenated with the sampled latent vector and then passed on to the decoder. The author claims that applying VS to the VAE model solves the bypass phenomenon, however, no evidence for this claim is provided.

The first published time-series anomaly detection approach based on VAE was LSTM-VAE \cite{park_multimodal_2018}. One of the contributions is its use of a dynamic prior $\mathcal{N}(\mu_p, 1)$, rather than a static one $\mathcal{N}(0, 1)$. In addition to that, they introduce a state-based threshold estimation method consisting of a support-vector regressor (SVR), which maps the latent distribution parameters $(\mu_{\textbf{z}}, \sigma_{\textbf{z}})$ to the resulting anomaly score using the validation data. Hence, the dynamic threshold can be obtained through Equation \ref{eq:svr}.
\begin{equation}
    \eta_t = \text{SVR}(\mu_{\textbf{z}, t}, \sigma_{\textbf{z}, t}) + c
    \label{eq:svr}
\end{equation}
where $c$ is a pre-defined constant to control sensitivity.

OmniAnomaly \cite{su_robust_2019} attempts to create a temporal connection between latent distributions by applying a linear Gaussian state space model to them. For the purpose of this paper, it is called (OmniA). Also, it concatenates the last gated recurrent unit (GRU) hidden state with the latent vector sampled in the previous time step. In addition to that, it uses planar normalising flow \cite{rezende_variational_2015} by applying $K$ transformations to the latent vector in order to approximate a non-Gaussian posterior, as shown in Equation \ref{eq:pnf}.
\begin{equation}
    f^k(\textbf{z}_t^{k-1}) = \textbf{u}\tanh(\textbf{w}\textbf{z}^{k-1}_t)+\textbf{b}
    \label{eq:pnf}
\end{equation}
where $\textbf{u}$, $\textbf{w}$ and $\textbf{b}$ are trainable parameters.

A simplified VAE architecture \cite{pereira_unsupervised_2019} based on BiLSTM layers is also proposed. For the purpose of our paper, it is called W-VAE. Unlike its predecessor \cite{pereira_unsupervised_2018}, it drops the attention mechanism but provides contributions elsewhere. It offers two strategies to detect anomalies based on the VAE outputs. The first involves clustering the space characterised by the mean parameter of the latent distribution into two clusters and labelling the larger one as normal. This strategy has a few weaknesses: it cannot be used in an operative environment as it requires some sort of history of test windows to form the clusters and it assumes that there are always anomalous samples present. The second strategy finds the Wasserstein similarity measure (hence the W in the name) between the latent mean space mapping of the test window in question and the respective mapping $i$ resulting from a representative data subset, such as the validation subset. Equation \ref{eq:wasserstein} shows how the Wasserstein similarity measure is computed
\begin{equation}
    W_i(\textbf{z}_{\text{test}}, \textbf{z}_{i}) = \|\mu_{\textbf{z}_\text{test}} - \mu_{\textbf{z}_{i}}\|_2^2 + \|\Sigma_{\textbf{z}_\text{test}}^{1/2} - \Sigma_{\textbf{z}_{i}}^{1/2}\|_F^2
    \label{eq:wasserstein}
\end{equation}
where the first term represents the $L2$-Norm between the mean distribution parameters resulting from the test window and each point of the representative subset. The second term represents the Frobenius norm between the covariance matrix resulting from the test window and each point of the representative subset.

SWCVAE \cite{chen_unsupervised_2020} is the first that applies convolutional neural networks (CNN) to VAE for multivariate time-series anomaly detection. Peculiarly, 2D CNN layers are used with the justification of being able to process the input both spatially and temporally. We, however, doubt the ability of the model to properly detect anomalies through spatial processing, as a kernel moving along the feature axis can only capture features adjacent to each other. To create a continuous anomaly score from windows they append the last value of each window to the previous one. For the purpose of this paper, this process is referred to as \textit{last-type reverse-windowing}.

SISVAE \cite{li_anomaly_2021} tries to improve the modelling robustness by the addition of a smoothing term in the loss function which contributes to the reduction of sudden changes in the reconstructed signal, making it less sensitive to noisy time steps.

As part of the VASP framework \cite{von_schleinitz_vasp_2021}, a variational autoencoder architecture is proposed to increase the robustness of time-series prediction when faced with anomalies. While the main contribution is attributed to the framework itself, not the VAE, it should be noted that during inference only the mean parameter of the latent distribution is passed to the decoder.

\section{Proposed approach}\label{sec:proposed_approach}
\begin{figure*}[t!]
    \centering
    \includegraphics[width=0.9\textwidth]{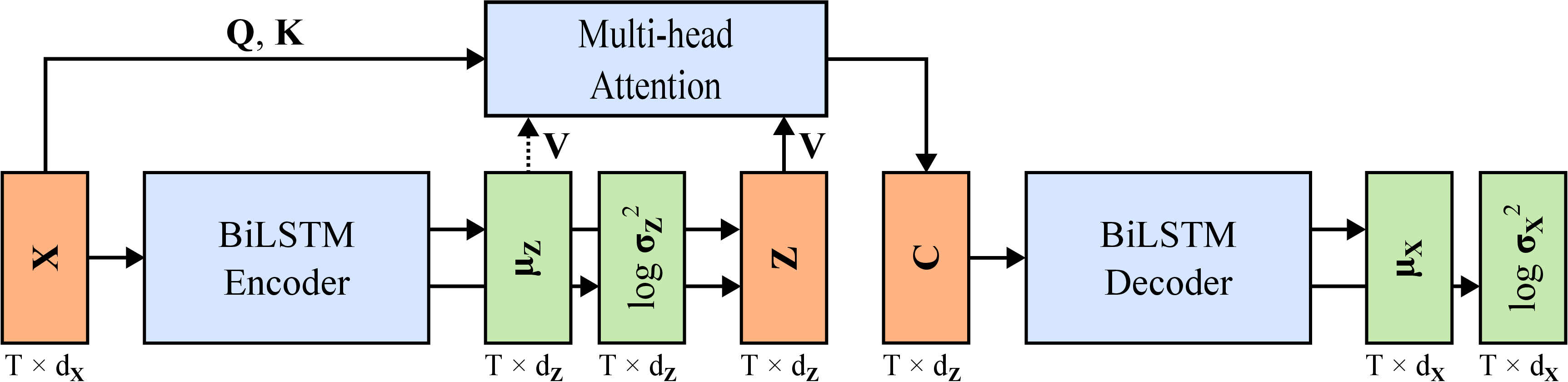}
    \caption{An illustration of the proposed MA-VAE model. Blue shapes designate trainable models, orange deterministic tensors and green distribution parameters. The shape of each tensor is designated below it. During training $\textbf{Z}$ is used as the value matrix, denoted by the solid arrow, whereas during inference $\boldsymbol{\mu}_\textbf{Z}$ is used as the value matrix, denoted by the traced arrow.} 
    \label{fig:MA-VAE_diagram_training}
\end{figure*}
\subsection{Overview}
To detect anomalies in multivariate time-series data, we propose a variational autoencoder architecture consisting of BiLSTM layers. The model architecture is illustrated in Figure \ref{fig:MA-VAE_diagram_training}. During training, the encoder $q_{\phi}$ maps multivariate input window $\textbf{X}$ to a temporal distribution with parameters $\boldsymbol{\mu}_\textbf{Z}$ and $\log\boldsymbol{\sigma}_\textbf{Z}^2$ in the forward pass, Equation \ref{eq:enc_fwd1}.
\begin{equation}
    (\boldsymbol{\mu}_\textbf{Z}, \log\boldsymbol{\sigma}_\textbf{Z}^2) = q_{\phi}(\textbf{X})
    \label{eq:enc_fwd1}
\end{equation}
Given the latent distribution parameters $\boldsymbol{\mu}_\textbf{Z}$ and $\log\boldsymbol{\sigma}_\textbf{Z}^2$, the latent matrix is sampled from the resulting distribution, as shown in Equation \ref{eq:enc_fwd2}.
\begin{equation}
    \textbf{Z} \sim \mathcal{N}(\boldsymbol{\mu}_\textbf{Z}, \log\boldsymbol{\sigma}_\textbf{Z}^2)
    \label{eq:enc_fwd2}
\end{equation}
Then, the input window $\textbf{X}$ is linearly transformed to obtain the query matrices $\textbf{Q}_i$ and key matrices $\textbf{K}_i$ for each head $i$ . Likewise, the sampled latent matrix $\textbf{Z}$ is also transformed to the value matrix $\textbf{V}_i$, as shown in Equation \ref{eq:att_fwd1}. 
\begin{equation}
    \textbf{Q}_i = \textbf{X}\textbf{W}^Q_i \quad \textbf{K}_i = \textbf{X}\textbf{W}^K_i \quad \textbf{V}_i = \textbf{Z}\textbf{W}^V_i
    \label{eq:att_fwd1}
\end{equation}
To output the context matrix $\textbf{C}_i$ for each head $i$, the softmax of the through $\sqrt{d_\textbf{K}}$ normalised query and key product is multiplied with the value matrix, Equation \ref{eq:att_fwd2}.
\begin{equation}
    \textbf{C}_i = \mathrm{Softmax}\left(\frac{\textbf{Q}_i\textbf{K}_i^T}{\sqrt{d_\textbf{K}}}\right)\textbf{V}_i
    \label{eq:att_fwd2}
\end{equation}
The final context matrix $\textbf{C}$ is the result of the linearly-transformed concatenation of each head-specific context matrix $\textbf{C}_i$, as expressed in Equation \ref{eq:att_fwd3}. 
\begin{equation}
    \textbf{C} = [\textbf{C}_1, ..., \textbf{C}_h]\textbf{W}^{O}
    \label{eq:att_fwd3}
\end{equation}
The decoder $p_{\theta}$ then maps the context matrix $\textbf{C}$ to an output distribution with parameters $\boldsymbol{\mu}_\textbf{X}$ and $\log\boldsymbol{\sigma}_\textbf{X}^2$, as shown in Equation \ref{eq:dec_fwd1}.
\begin{equation}
    (\boldsymbol{\mu}_\textbf{X}, \log\boldsymbol{\sigma}_\textbf{X}^2) = p_{\theta}(\textbf{C})
    \label{eq:dec_fwd1}
\end{equation}

\subsection{Inference Mode}
Despite the generative capabilities of VAE, MA-VAE does not leverage generation for anomaly detection. Rather than sampling a latent matrix as shown in Equation \ref{eq:enc_fwd2} during inference, sampling is disabled and only $\boldsymbol{\mu}_\textbf{Z}$ is taken as the input for the multi-head attention mechanism, like in \cite{von_schleinitz_vasp_2021}. Equation \ref{eq:enc_fwd2} in the forward pass, therefore, is replaced by Equation \ref{eq:inference}.
\begin{equation}
    \textbf{Z} = \boldsymbol{\mu}_\textbf{Z}
    \label{eq:inference}
\end{equation}
This not only accelerates inference by eliminating the sampling process but is also empirically found to be a good approximation of an averaged latent matrix if it were sampled several times like in \cite{pereira_unsupervised_2018}. The MA-VAE layout during inference is shown in Figure \ref{fig:MA-VAE_diagram_training}, where the traced arrow designates the information flow from the encoder to the MA mechanism.

\subsection{Threshold Estimation Method}
Anomalies are by definition very rare events, hence an ideal anomaly detector only flags measurements very rarely but accurately. Test bench engineers prefer an algorithm that only flags a sequence it is sure is an anomaly, in other words, an algorithm that outputs very few to no false positives. A high false positive count would lead to a lot of stoppages and therefore lost testing time and additional cost. Of course, the vast majority of measurements evaluated will be normal and hence it is paramount to classify them correctly, naturally leading to a high precision value. Also, there is no automatic evaluation methodology currently running at test benches, other than rudimentary rule-based methods, therefore a solution that plugs into the existing system that automatically detects \textit{some} or \textit{most} anomalies undetectable by rules is already a gain. To achieve this, the threshold $\tau$ is set as the maximum log probability observed when the model is fed with validation data.

\subsection{Bypass Phenomenon}
VAE, when combined with an attention mechanism, can exhibit a behaviour called the bypass phenomenon \cite{bahuleyan_variational_2018}. When the bypass phenomenon happens the latent path between encoder and decoder is ignored and information flow occurs mostly or exclusively through the attention mechanism, as it has deterministic access to the encoder hidden states and therefore avoids regularisation through the $D_{\text{KL}}$ term. In an attempt to avoid this, \cite{bahuleyan_variational_2018} propose variational attention, which, like the VAE, maps the input to a distribution rather than a deterministic vector. Applied to natural language processing, \cite{bahuleyan_variational_2018} demonstrate that this leads to a diversified generated portfolio of sentences, indicating alleviation of the bypassing phenomenon. As previously mentioned, only \cite{pereira_unsupervised_2018} applies this insight in the anomaly detection domain, however, they do not present any proof that it alleviates the bypass phenomenon in their work. MA-VAE on the other hand, cannot suffer from the bypass phenomenon in the sense that information flow ignores the latent variational path between encoder and decoder since the MA mechanism requires the value matrix \textbf{V} from the encoder to output the context matrix. Assuming the bypass phenomenon also applies to a case where information flow ignores the attention mechanism, one could claim that MA-VAE is not immune. To disprove this claim, the attention mechanism is removed from the model in an ablation study to see if anomaly detection performance remains the same. In this case, $\textbf{V}$ is instead directly input into the decoder. If it drops, it is evidence of the contribution of the attention mechanism to the model performance and hence is not bypassed. The results for this ablation study are shown and discussed in Section \ref{sec:results}.

\subsection{Impact of Seed Choice}
Given the stochastic nature of the VAE, the chosen seed can impact the anomaly detection performance as it can lead to a different local minimum during training. To investigate the impact the seed choice has on model training, MA-VAE is trained on three different seeds, the respective results are also shown in Section \ref{sec:results}. 

\subsection{Reverse-window Process}
Since the model is trained to reconstruct fixed-length windows, the same applies during inference. However, to decide whether a given measurement sequence $\mathcal{S} \in \mathbb{R}^{T \times d_\textbf{X}}$ is anomalous, a continuous reconstruction of the measurement is required. The easiest way to do so would be to window the input measurement using a shift of $1$, input the windows into the model and chain the last time step from each output window to obtain a continuous sequence \cite{chen_unsupervised_2020}. Considering the BiLSTM nature of the encoder and decoder, the first and last time steps of a window can only be computed given the states from one direction, making these values, in theory, less accurate, however. To overcome this, we propose averaging matching time steps in overlapping windows, which is called \textit{mean-type} reverse-window method. This is done by pre-allocating an array with NaN values, filling it, and taking the mean for each time step while ignoring the NaN values. This process and the general anomaly detection process are described in Algorithm \ref{alg:anomaly_detection}.
This reverse-window process is done for the mean and \textit{variance} parameters of the output distribution, then the variance is converted to \textit{standard deviation} since two distributions cannot be combined by averaging the standard deviations. With a continuous mean and standard deviation, the continuous negative log probability, i.e. the anomaly score $s$, is computed for the respective measurement. A comparison between the mean, last and first reverse-window process is provided in Section \ref{sec:results}.

\begin{algorithm}[t!]
\caption{Anomaly Detection Process}\label{alg:anomaly_detection}
\begin{algorithmic}
\Input $\text{Sequence  } \mathcal{S} \in \mathbb{R}^{T \times d_\textbf{X}}, \quad  \text{Threshold  } \tau$ 
\Result $\text{Label  }l$
\State $n_{\text{windows}} \gets T-W+1$

\State $\boldsymbol{\mu}_{\textbf{X},\text{temp}} \gets \text{zeros}(n_{\text{windows}}, T, d_\textbf{X}) + \text{NaN}$

\State $\boldsymbol{\sigma}_{\textbf{X},\text{temp}}^{2} \gets \text{zeros}(n_{\text{windows}}, T, d_\textbf{X}) + \text{NaN}$

\For{$i=1 \to n_{\text{windows}}$}
    \State $\textbf{X} \gets \mathcal{S}[i:W+i]$ \;
    
    \State $(\boldsymbol{\mu}_{\textbf{Z}}, \log \boldsymbol{\sigma}_{\textbf{Z}}^2) \gets q_{\phi}(\textbf{X})$\;   
    
    \State $\textbf{C} \gets \text{MA}(\textbf{X}, \textbf{X}, \boldsymbol{\mu}_{\textbf{Z}})$\; 
    
    \State $(\boldsymbol{\mu}_{\textbf{X}}, \log \boldsymbol{\sigma}_{\textbf{X}}^2) \gets p_{\theta}(\textbf{C})$\; 

    \State $\boldsymbol{\mu}_{\textbf{X},\text{temp}}[i, i: i+W] \gets \boldsymbol{\mu}_{\textbf{X}}$ 
    
    \State $\boldsymbol{\sigma}_{\textbf{X},\text{temp}}^{2}[i, i: i+W] \gets \boldsymbol{\sigma}_{\textbf{X}}^2$ 

\EndFor

\State $\boldsymbol{\mu}_{\textbf{X},\text{seq}} \gets \text{nanmean}(\boldsymbol{\mu}_{\textbf{X},\text{temp}})$

\State $\boldsymbol{\sigma}_{\textbf{X},\text{seq}}^2 \gets \text{nanmean}(\boldsymbol{\sigma}_{\textbf{X},\text{temp}}^2)$
    
\State $s \gets -\log p(\textbf{X} \vert \boldsymbol{\mu}_{\textbf{X},\text{seq}}, \boldsymbol{\sigma}_{\textbf{X},\text{seq}})$\;

\State $l \gets \text{max}(s) > \tau$\; 

\end{algorithmic}
\end{algorithm}

\section{Results}\label{sec:results}
\subsection{Setup}
The encoder and decoder both consist of two BiLSTM layers, with the outer ones having 512 hidden- and cell-state sizes and the inner ones 256. All other parameters are left as the default in the TensorFlow API. 

During training only, input windows are corrupted using Gaussian noise using $0.01$ standard deviation to increase robustness to noise.

Key factors that are investigated in Section \ref{sec:results} are given a default value which applies to all experiments unless otherwise specified. These factors are training/validation subset size, which is set to $512$h, seed choice, which has been kept at $1$, reverse-window method, where the mean-type is used, the latent dimension size, which is set to $d_{\textbf{Z}}=16$ and the MA mechanism, which is set up as proposed in \cite{vaswani_attention_2017} with a head count of $h=8$ and a key dimension size $d_\textbf{K}=\lfloor d_\textbf{X}/h \rfloor=1$.

The optimiser used is the AMSGrad optimiser with the default parameters in the TensorFlow API.

Cyclical $D_{\text{KL}}$ annealing \cite{fu_cyclical_2019} is applied to the training of MA-VAE, to avoid the $D_{\text{KL}}$ vanishing problem. The $D_{\text{KL}}$ vanishing problem occurs when regularisation is too strong at the beginning of training, i.e. the Kullback-Leibler divergence term has a larger magnitude in relation to the reconstruction term. Cyclical $D_{\text{KL}}$ annealing allows the model to weigh the Kullback-Leibler divergence lower than the reconstruction term in a cyclical manner through a weight $\beta$. This callback is configured with a grace period of 25 epochs, where $\beta$ is linearly increased from $0$ to $10^{-8}$. After the grace period, $\beta$ is set to $10^{-8}$ and is gradually increased linearly to $10^{-2}$ throughout the following $25$ epochs, representing one loss cycle. This loss cycle is repeated until the training stops.

All priors in this work are set as standard Gaussian distributions, i.e. $p = \mathcal{N}(0,1)$.

To prevent overfitting, early stopping is implemented. It works by monitoring the log probability component of the validation loss during training and stopping if it does not improve for $250$ epochs. Logically, the model weights at the lowest log probability validation loss are saved.

Training is done on a workstation configured with an NVIDIA RTX A6000 GPU. The library used for model training is TensorFlow 2.10.1 on Python 3.10 on Windows 10 Enterprise LTSC version 21H2.

The results provided are given in the form of the calibrated and uncalibrated anomaly detection performance, i.e. with and without consideration of threshold $\tau$, respectively. Recall that the threshold used is the absolute maximum negative log probability obtained from the validation set. Calibrated metrics are the precision, recall and $F_1$ score. Precision $P$ represents the ratio between correctly identified anomalies (true positives) and all positives (true and false), shown in Equation \ref{eq:precision_recall}, recall $R$ represents the ratio between true positives and all anomalies, shown in Equation \ref{eq:precision_recall}, and $F_1$ score represents the harmonic mean of the precision and recall, shown in Equation \ref{eq:f1}. The underlying metrics used to calculate all of the below are the true positives ($TP$), false negatives ($FN$) and false positives ($FP$).
\begin{equation}
    P=\frac{TP}{TP+FP} \quad R=\frac{TP}{TP+FN}
    \label{eq:precision_recall}
\end{equation}

\begin{equation}
    F_1=\frac{TP}{TP + 0.5 * (FP + FN)} = 2*\frac{P* R}{P+R}
    \label{eq:f1}
\end{equation}
The theoretical maximum $F_1$ score, $F_{1,\text{best}}$, is also provided to aid discussion. This represents the best possible score achievable by the approach if the ideal threshold were known, i.e. the point on the precision-recall curve that comes closest to the $P=R=1$ point, though, in reality, this value is not observable and hence \textit{cannot} be obtained in an unsupervised manner.

The uncalibrated anomaly detection performance, i.e. the performance for a range of thresholds, each $0.1$ apart, is represented by the area under the continuous precision-recall curve $A_\text{PRC}^\text{cont}$, Equation \ref{eq:cont_aucpr}.
\begin{equation}
    A_\text{PRC}^\text{cont}=\int_0^1 P \,dR 
    \label{eq:cont_aucpr}
\end{equation}
As the integral cannot be computed for the continuous function, the area under the discrete precision-recall curve $A_\text{PRC}^\text{disc}$ is used which is done using the trapezoidal rule, Equation \ref{eq:disc_aucpr}.
\begin{equation}
    A_\text{PRC}^\text{disc}=\sum^{N}_{k=1}\frac{f(R_{k-1})+f(R_{k})}{2}\Delta R_k
    \label{eq:disc_aucpr}
\end{equation}
where $N$ is the number of discrete sub-intervals, $k$ the index of sub-intervals and $\Delta R_k$ the sub-intervals length at index $k$. Precision is a function of recall, i.e. $P=f(R)$.

\subsection{Ablation Study}
MA-VAE is tested without the MA mechanism and with a direct connection from the encoder to the decoder to observe whether it impacts results.

\begin{figure}[h]
    \centering
    \includegraphics[width=0.4\textwidth]{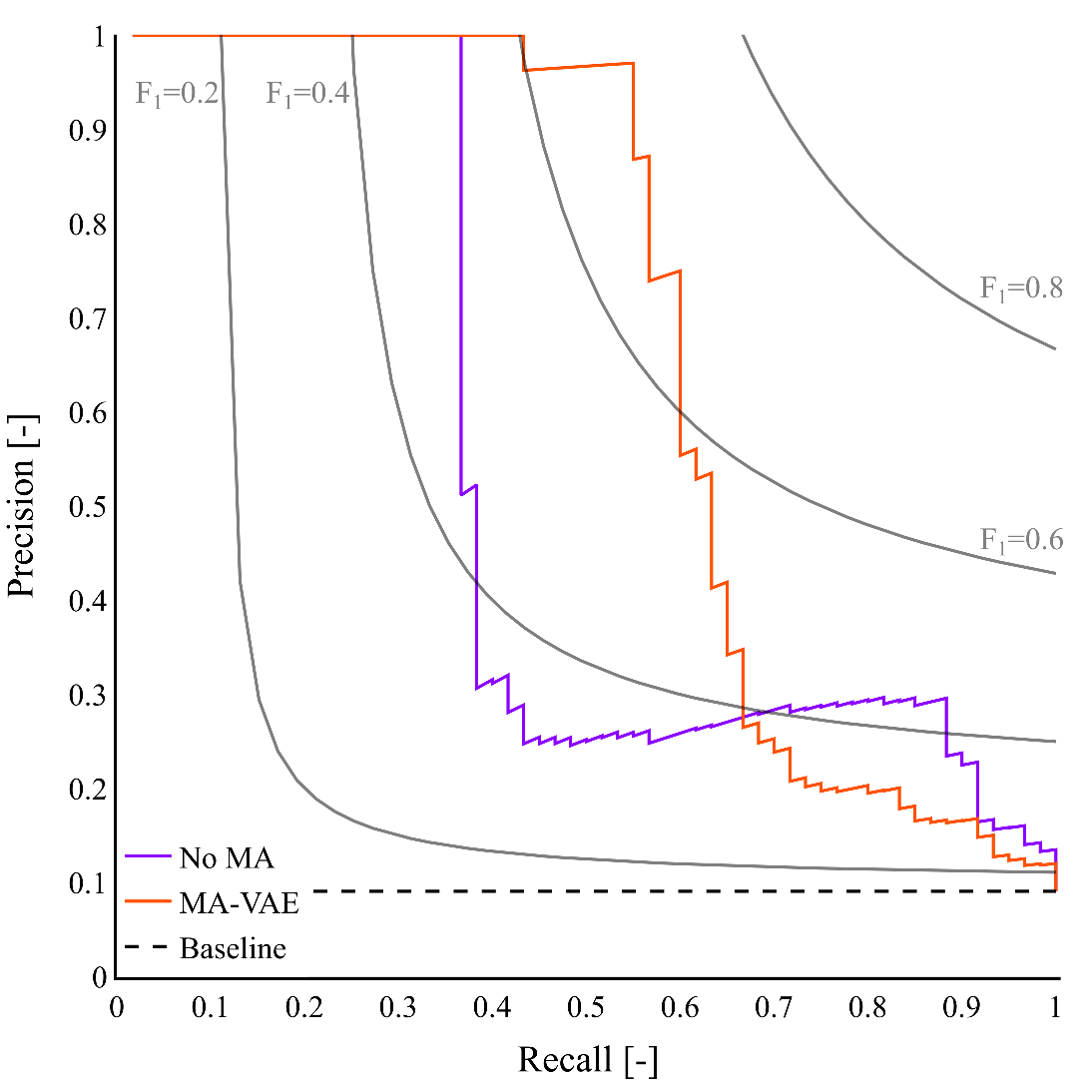}
    \caption{Precision-recall curves for the model variation without MA and MA-VAE.}
    \label{fig:aucpr_ablation}
\end{figure}

The anomaly detection performance of MA-VAE and its counterpart without MA, henceforth referred to as \textit{No MA} model, are shown in Table \ref{tab:ablation}. While the precision value of the No MA model is slightly higher than the MA-VAE, the recall value on the other hand is much lower. Overall, MA-VAE has a higher $F_1$ score, as well as a higher theoretical maximum $F_1$ score, although both values are so close enough to each other that one could claim the threshold is near ideal. The uncalibrated performance is also higher in the case of the MA-VAE, as evident in the precision-recall plot in Figure \ref{fig:aucpr_ablation}. Interestingly, MA-VAE may feature a lower precision value for the chosen unsupervised threshold but has the potential to have a higher maximum recall at $P=1$.

The results hence point towards an improvement brought about by the addition of the MA mechanism and therefore the bypass phenomenon can be ruled out.

\begin{table}[h]
\centering
\caption{Precision $P$, recall $R$, $F_1$ score, theoretical best $F_1$ score $F_{1,\text{best}}$ and area under the precision-recall curve $A_{PRC}$ results for the model variant without the MA mechanism and MA-VAE. The best values for each metric are given in \textbf{bold}.}\label{tab:ablation}
\begin{tabular}{cccccc}\hline
Model    & $P$  & $R$  & $F_1$ & $F_{1,\text{best}}$ & $A_{\text{PRC}}$ \\ \hline \hline
No MA    & \textbf{1.00} & 0.35 & 0.52  & 0.54         & 0.52                    \\
MA-VAE   & 0.92 & \textbf{0.55} & \textbf{0.69}  & \textbf{0.70}         & \textbf{0.66}                    \\ \hline
\end{tabular}
\end{table}

\subsection{Data Set Size Requirements}
To evaluate how much data is required to train MA-VAE to a point of adequate anomaly detection performance, it has been trained with $1$h, $8$h, $64$h, and $512$h worth of dynamic testing data.

\begin{figure}[h]
    \centering
    \includegraphics[width=0.4\textwidth]{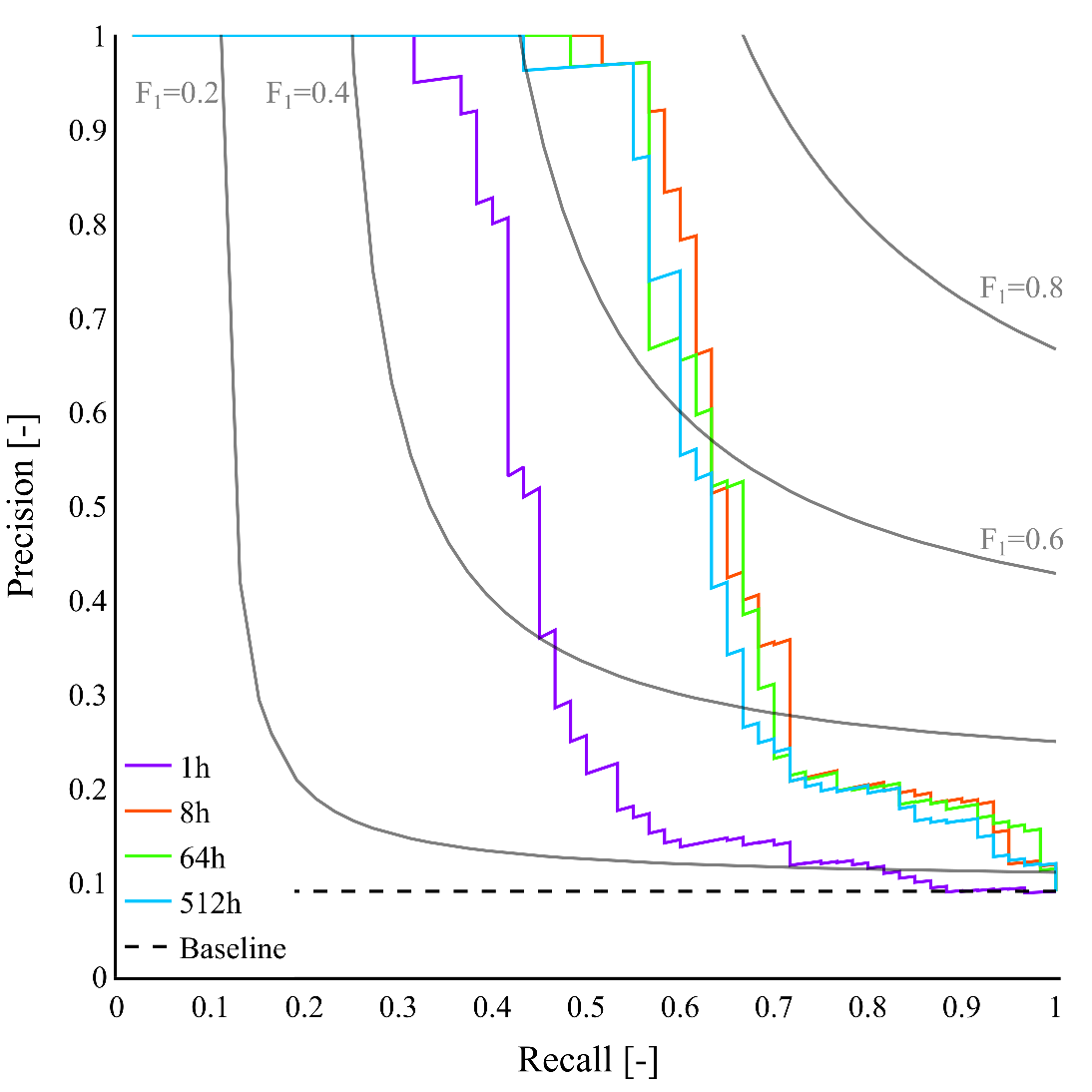}
    \caption{Precision-recall curves for the model trained on different training/validation subset sizes.}
    \label{fig:aucpr_dataset}
\end{figure}

The results for this experiment are presented in Table \ref{tab:dataset}. On the one hand, as the training/validation subset increases in size, the precision value improves, with the largest jump occurring when the dynamic testing time goes from $1$h to $8$h. The recall value on the other hand decreases as the subset grows. This can be attributed to the fact that smaller subset sizes lead to a small validation set and therefore less data to obtain a threshold from. With a limited amount of data to obtain a threshold from, it is more difficult to get a representative error distribution, leading to a threshold that is very small and hence marks most anomalies correctly but also leads to a lot of false positives. $F_1$ score reaches a point of diminishing returns with the $8$h subset onwards, this can also be observed in the case of the theoretical maximum $F_1$ score, $F_{1,\text{best}}$, as well as in the $A_{\text{PRC}}$ value, further supported by the precision-recall plot in Figure \ref{fig:aucpr_dataset}. Lastly, the $F_1$ score seems to approach the $F_{1,\text{best}}$ score as the subset grows, also backing the fact that with a small subset size, a good threshold cannot easily be obtained.
\begin{table}[h]
\centering
\caption{Precision $P$, recall $R$, $F_1$ score, threoretical best $F_1$ score $F_{1,\text{best}}$ and area under the precision-recall curve $A_{PRC}$ results for the different training/validation subset sizes. The best values for each metric are given in \textbf{bold}.}\label{tab:dataset}
\begin{tabular}{cccccc}\hline
Size   & $P$  & $R$  & $F_1$ & $F_{1,\text{best}}$ & $A_{\text{PRC}}$ \\ \hline \hline
$1$h   & 0.09 & \textbf{0.88} & 0.17  & 0.55         & 0.49                    \\
$8$h   & 0.66 & 0.63 & 0.64  & \textbf{0.72}         & \textbf{0.69}                    \\
$64$h  & 0.71 & 0.57 & 0.63  & 0.69         & 0.68                    \\
$512$h & \textbf{0.92} & 0.55 & \textbf{0.69}  & 0.70         & 0.66                    \\ \hline
\end{tabular}
\end{table}

Therefore, for application at the test bench, the largest subset size is desirable due to the higher precision value and a \textit{closer-to-ideal} threshold value.

\subsection{Impact of Seed Choice}
To illustrate the impact it has on the performance metrics, they are presented using three different seeds. 

\begin{figure}[h]
    \centering
    \includegraphics[width=0.4\textwidth]{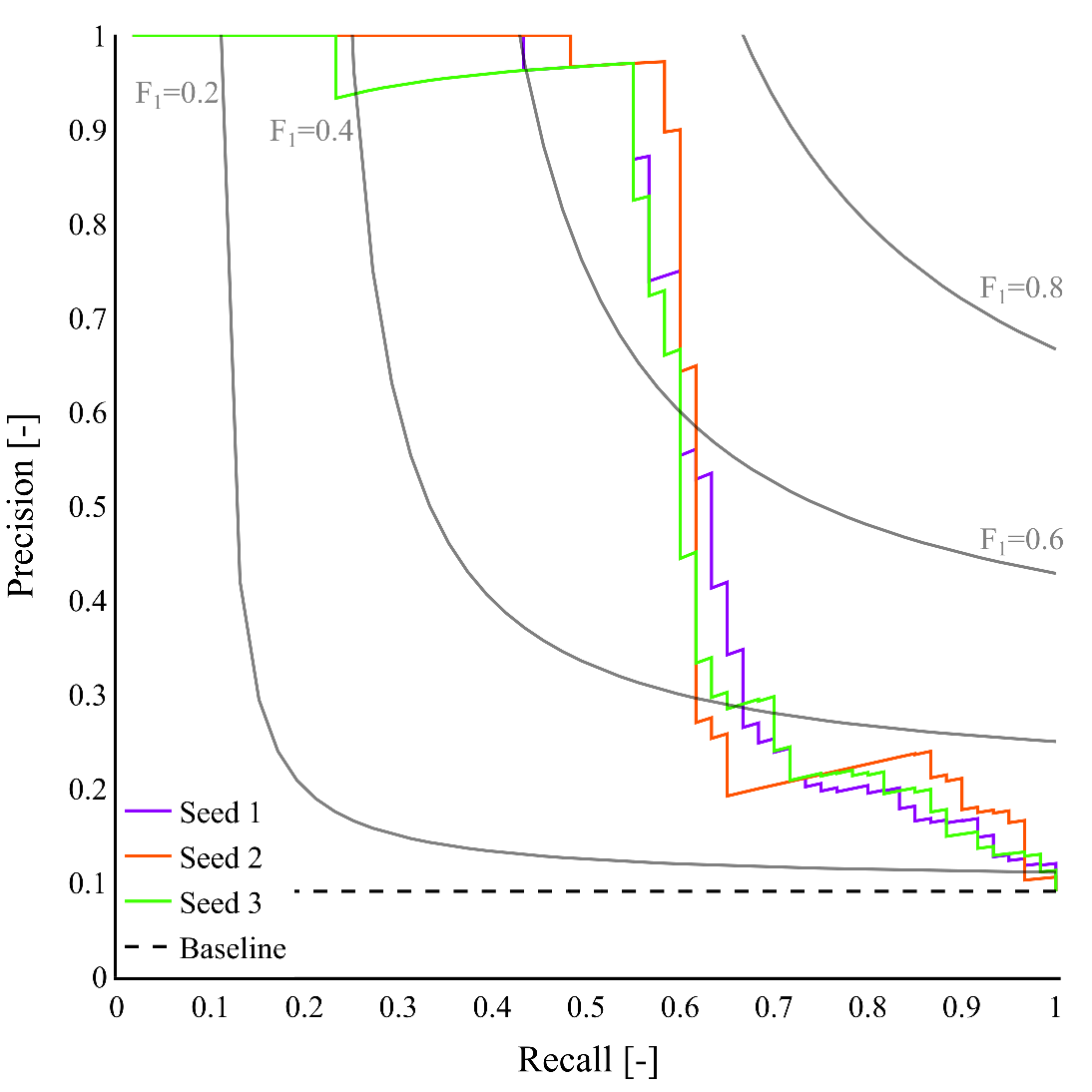}
    \caption{Precision-recall curves for the model trained on different seeds.}
    \label{fig:aucpr_seed}
\end{figure}

Table \ref{tab:seed} shows that while the precision values are roughly in the same range for all seeds, the recall values vary more significantly, which also reflects on the $F_1$ score. However, by inspecting the $F_{1,\text{best}}$ and $A_{\text{PRC}}$ values it becomes clear that the seeds are not as far apart as the recall value suggests and that the issue may lie with the threshold choice. Figure \ref{fig:aucpr_seed} further supports this, as all lines have roughly the same path, with the exception of seed $3$ at very high precision values. The plot clearly shows that a more suitable (lower) threshold would lead to seed $3$ having a comparable recall value to the other seeds while maintaining high precision. 

Some differences can be observed between the seeds, especially in the recall values, however, this can be attributed to the unsupervised threshold choice. 

\begin{table}[h]
\centering
\caption{Precision $P$, recall $R$, $F_1$ score, threoretical best $F_1$ score $F_{1,\text{best}}$ and area under the precision-recall curve $A_{PRC}$ results for the different seeds. The best values for each metric are given in \textbf{bold}.}\label{tab:seed}
\begin{tabular}{cccccc}\hline
Seed & $P$  & $R$  & $F_1$ & $F_{1,\text{best}}$ & $A_{\text{PRC}}$ \\ \hline \hline
1    & 0.92 & 0.55 & 0.69  & 0.70                & 0.66             \\
2    & 0.90 & \textbf{0.60} & \textbf{0.72}  & \textbf{0.73}                & \textbf{0.67}             \\
3    & \textbf{0.96} & 0.40 & 0.56  & 0.70                & 0.64             \\ \hline
\end{tabular}
\end{table}

\subsection{Reverse-window Process}
To investigate the effect of the mean-type reverse-window method, it is compared with the first-type and last-type methods where the first and last values of each window are carried over, respectively.

\begin{figure}[t]
    \centering
    \includegraphics[width=0.4\textwidth]{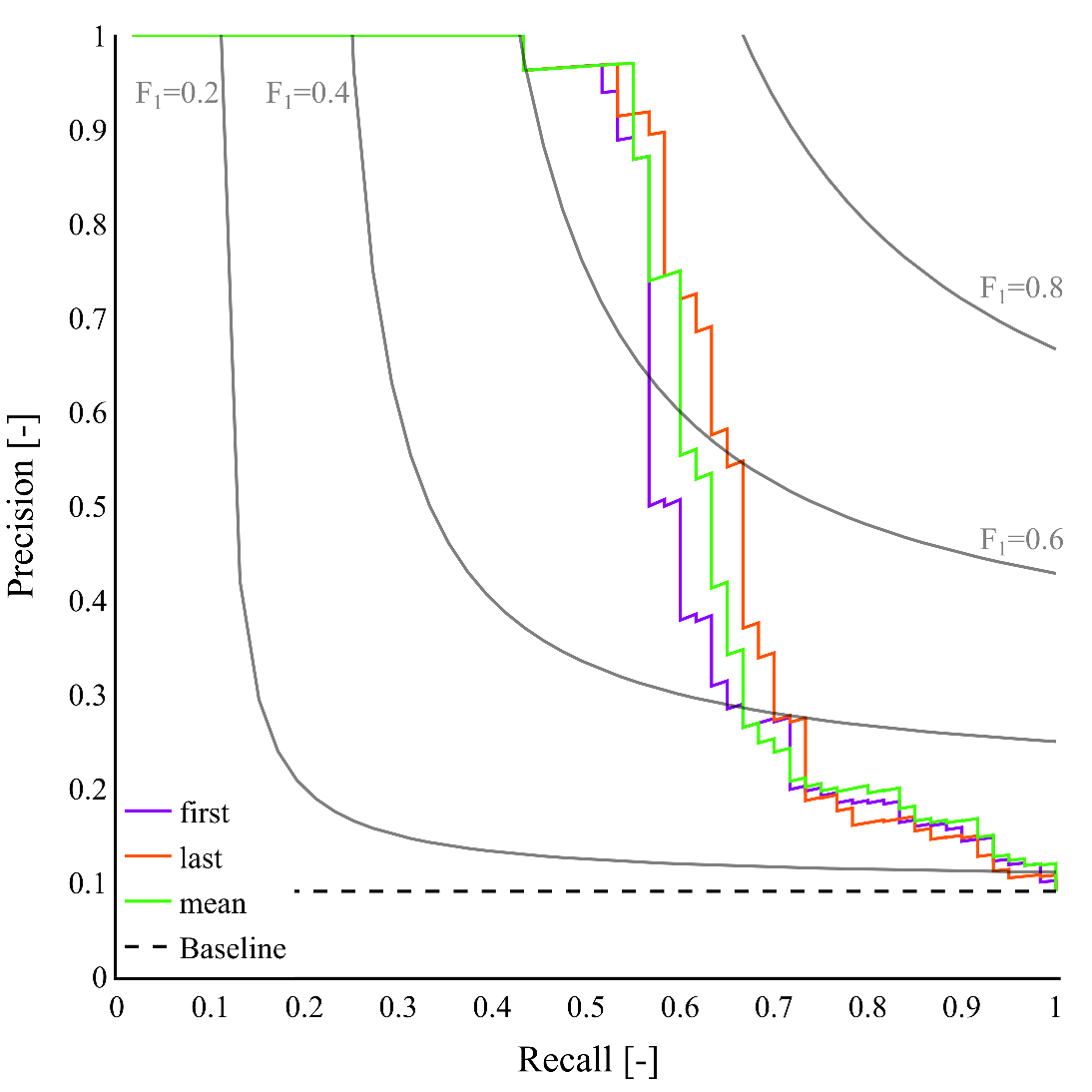}
    \caption{Precision-recall curves for different reverse-window methods.}
    \label{fig:aucpr_reverse}
\end{figure}

The results in this subsection, Table \ref{tab:reverse} and Figure \ref{fig:aucpr_reverse}, tell a similar story to the previous subsection. The metrics independent of the chosen threshold are very similar regardless of the reverse-window method, implying that they are comparable and that any differences in the calibrated metrics can be attributed to the chosen threshold. The mean-type reverse-window method results in a higher computational load, though negligible. For a rather long sequence of $4000$ time steps, i.e. around $33$ minutes long, the mean-type method only takes around $2$ seconds longer. One source of delay that can appear, however, is during \textit{online} anomaly detection. An online anomaly detection algorithm is defined as an algorithm which evaluates the sequence as it is being recorded. To obtain time step $t$ using the mean-type (or the first-type) you have to wait for time step $t+W$ while $t<W$. This translates to a delay of around $2$ minutes in the real world, given the chosen window size. If the evaluation is done offline, i.e. when $t=W$, then this delay is eliminated since the last value does not have other overlapping values to compute the mean.

\begin{table}[t]
\centering
\caption{Precision $P$, recall $R$, $F_1$ score, threoretical best $F_1$ score $F_{1,\text{best}}$ and area under the precision-recall curve $A_{PRC}$ results for the different reverse-window types. The best values for each metric are given in \textbf{bold}.}\label{tab:reverse}
\begin{tabular}{cccccc}\hline
Type    & $P$  & $R$  & $F_1$ & $F_{1,\text{best}}$ & $A_{\text{PRC}}$ \\ \hline \hline
first   & \textbf{0.97} & 0.48 & 0.64  & 0.69                & 0.64             \\
last    & 0.88 & \textbf{0.58} & \textbf{0.71}  & \textbf{0.71}                & \textbf{0.67}             \\
mean    & 0.92 & 0.55 & 0.69  & 0.70                & 0.66             \\ \hline
\end{tabular}
\end{table}

\subsection{Hyperparameter Optimisation}
As part of the hyperparameter optimisation of MA-VAE, a list of latent dimension sizes $d_\textbf{Z}$ in combination with a list of key dimension sizes $d_\textbf{K}$ is tested. Despite the larger learning capacity associated with a higher $d_\textbf{K}$, the concatenation is always transformed to a matrix of size $d_\textbf{O}=d_\textbf{Z}$. 
For the two variables, values of 1, 4, 16, and 64 are tested.

\begin{figure}[t]
    \centering
    \includegraphics[width=0.4\textwidth]{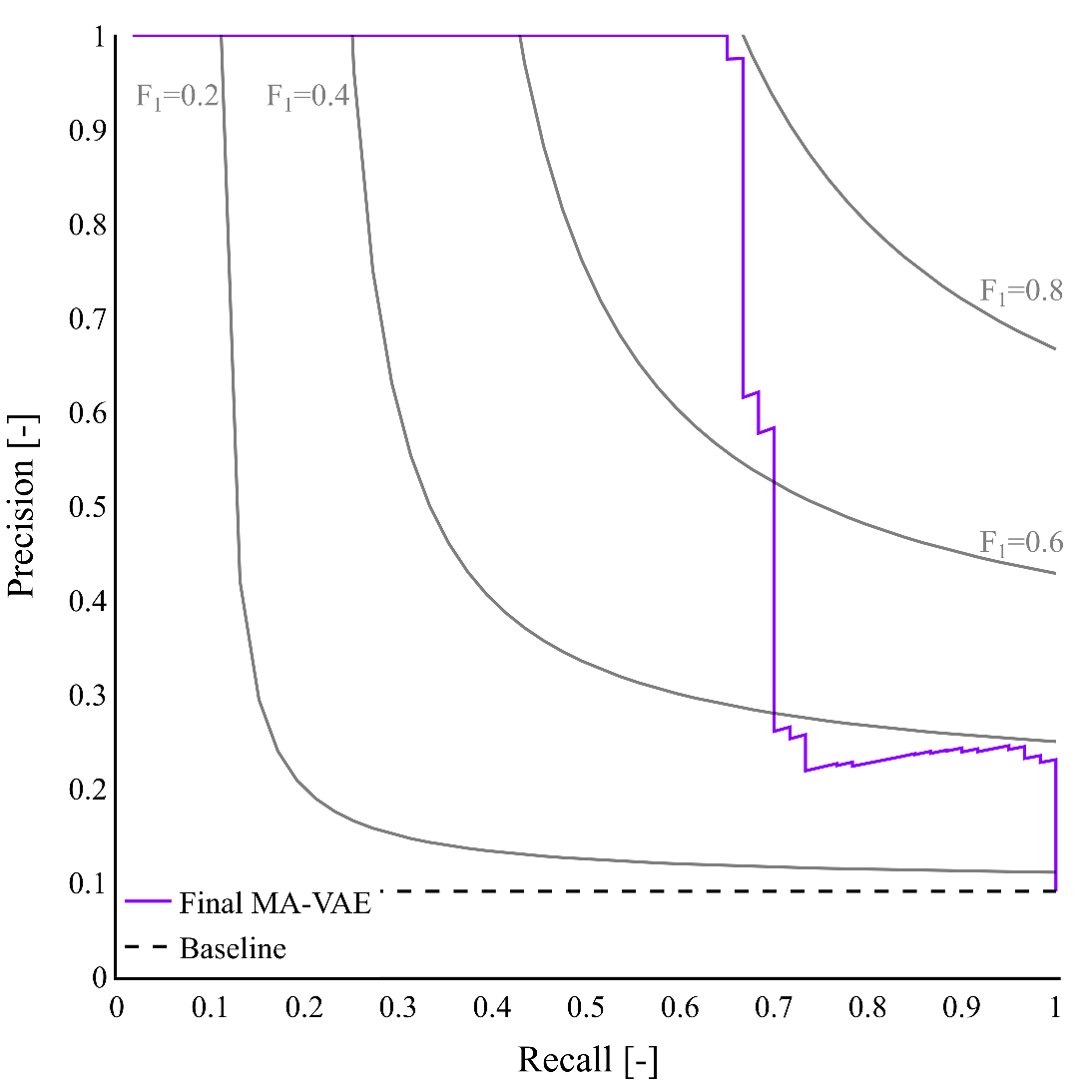}
    \caption{Precision-recall curve for the final MA-VAE.}
    \label{fig:hyperparameter}
\end{figure}

\begin{figure}[t]
    \centering
    \includegraphics[width=0.4\textwidth]{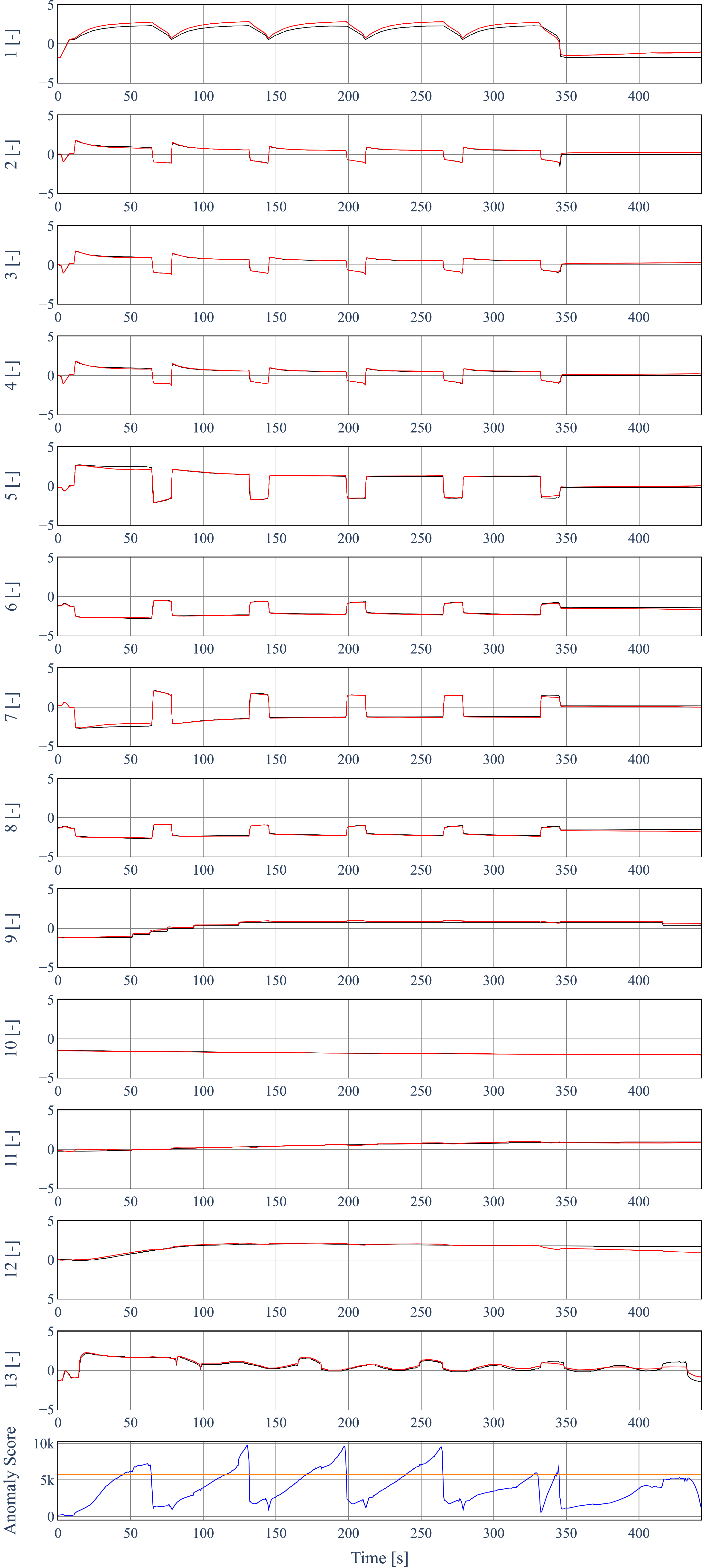}
    \caption{Wheel diameter anomaly plotted in black and the output distribution in red, as well as anomaly score plotted in blue and the threshold as a straight line in orange.}\label{fig:recon_wheeldiameter}
\end{figure}

The best result is achieved with $d_\textbf{Z}=d_\textbf{K}=64$. Given that they are the respective highest values of $d_\textbf{Z}$ and $d_\textbf{K}$, even higher values should be experimented with in the future, though they will lead to higher model complexity and training/inference time. The attention head count $h$ was also experimented with using the same range of values as for $d_\textbf{Z}$ and $d_\textbf{K}$, however, none performed better than the $h=8$ configuration. The results are presented in Table \ref{tab:hyperparameter}, the corresponding precision-recall plot is shown in Figure \ref{fig:hyperparameter}. $91\%$ of the sequences marked as normal were actually normal and $67\%$ of the total number of anomalous sequences in the test set were detected. One example of the anomalous cycles and the respective reconstructions is plotted in Figure \ref{fig:recon_wheeldiameter}. 
\begin{table}[t]
\caption{Precision $P$, recall $R$, $F_1$ score, threoretical best $F_1$ score $F_{1,\text{best}}$ and area under the precision-recall curve $A_{PRC}$ result for the best $d_\textbf{Z}$, $d_\textbf{K}$ and $h$ values.}\label{tab:hyperparameter}
\centering
\begin{tabular}{cccccccc} \hline
$d_\textbf{Z}$ & $d_\textbf{K}$ & $h$ & $P$  & $R$  & $F_1$ & $F_{1,\text{best}}$ & $A_{\text{PRC}}$ \\ \hline \hline
64             & 64           & 8 & 0.91 & 0.67 & 0.77  & 0.79                & 0.74             \\  \hline
\end{tabular}
\end{table}

\subsection{Benchmarking}
Of course, MA-VAE is not the first model proposed for time-series anomaly detection. To underline its anomaly detection performance, it is compared with a series of other models based on variational autoencoders. The chosen subset of models is based on the work discussed in Section \ref{sec:related_work} which either linked source code or contained enough information for implementation. The models are implemented using hyperparameters specified in their respective publications. To even the playing field, the models are trained on the $512$h subset with early stopping, which is parametrised equally across all models. The anomaly detection process specified in Algorithm \ref{alg:anomaly_detection} is also applied to all models, along with the threshold estimation method. The results can be seen in Table \ref{tab:benchmark}.

\begin{table}[b]
\caption{Precision $P$, recall $R$, $F_1$ score, threoretical best $F_1$ score $F_{1,\text{best}}$ and area under the precision-recall curve $A_{PRC}$ results for competing models and MA-VAE (Ours). The best values for each metric are given in \textbf{bold}.}\label{tab:benchmark}
\centering
\begin{tabular}{cccccc}
\hline
Model           & $P$  & $R$  & $F_1$ & $F_{1,\text{best}}$ & $A_{\text{PRC}}$ \\ \hline\hline
VS-VAE          & \textbf{1.00} & 0.33 & 0.50  & 0.56                & 0.51             \\ 
W-VAE           & \textbf{1.00} & 0.30 & 0.46  & 0.46                & 0.41             \\ 
OmniA           & 0.96 & 0.37 & 0.53  & 0.58                & 0.53             \\ 
SISVAE          & \textbf{1.00} & 0.30 & 0.46  & 0.50                & 0.51             \\ 
MA-VAE          & 0.91 & \textbf{0.67} & \textbf{0.77}  & \textbf{0.79}                & \textbf{0.74}             \\ \hline
\end{tabular}
\end{table}

\noindent As is evident, MA-VAE outperforms all other models in every metric, except for precision. As stated in Section \ref{sec:proposed_approach} a high precision figure is important in this type of powertrain testing, however, the reduced precision is still considered tolerable. Also, it comes at the benefit of a much higher recall figure, which is reflected in the superior $F_1$ figure. Furthermore, the $F_{1,\text{best}}$ figure, which is obtained at $P=0.98$ and $R=0.67$, suggests that MA-VAE has the potential to achieve even higher precision without sacrificing recall if the threshold were optimised. The higher $A_{PRC}$ also shows that MA-VAE has a higher range of thresholds at which it performs well.

\section{Conclusion and Outlook}\label{sec:conclusion}
In this paper, a multi-head attention variational autoencoder (MA-VAE) for anomaly detection in automotive testing is proposed. It not only features an attention configuration that avoids the bypass phenomenon but also introduces a novel method of remapping windows to whole sequences.
A number of experiments are conducted to demonstrate the anomaly detection performance of the model, as well as to underline the benefits of key aspects introduced with the model.

From the results obtained, MA-VAE clearly benefits from the MA mechanism, indicating the avoidance of the bypass phenomenon. Moreover, the proposed approach only requires a small training/validation subset size but fails to obtain a suitable threshold, as with increasing subset size only the calibrated anomaly detection performance increases. Training with different seeds also is shown to have little impact on the anomaly detection metrics, provided the threshold is chosen suitably, further underlining the previous point. Moreover, mean-type reverse windowing fails to significantly outperform its first-type and last-type counterparts, while introducing additional lag if it is applied to online anomaly detection. Lastly, the hyperparameter optimisation revealed that the MA-VAE variant with the largest latent dimension and attention key dimension resulted in the best anomaly detection performance. It is only $9\%$ of the time wrong when an anomaly is flagged and manages to discover $67\%$ of the anomalies present in the test data set. Also, it outperforms all other competing models it is compared with.

In the future, a method of threshold choice involving active learning will be investigated, which can use user feedback to hone in on a better threshold. Also, MA-VAE is set to be tested in the context of online anomaly detection, i.e. during the driving cycle measurement.

\bibliographystyle{apalike}
{\small
\bibliography{references}}

\end{document}